\documentclass{article}

    \PassOptionsToPackage{numbers, compress}{natbib}

\usepackage[preprint]{neurips_2020}

\usepackage[utf8]{inputenc} 
\usepackage[T1]{fontenc}    
\usepackage{hyperref}       
\usepackage{url}            
\usepackage{booktabs}       
\usepackage{amsfonts}       
\usepackage{nicefrac}       
\usepackage{microtype}      

\usepackage{multirow}

\usepackage{wrapfig}
\usepackage{epsfig}
\usepackage{graphicx}
\usepackage{amsmath}
\usepackage{amssymb}

\usepackage{makecell}
\usepackage{caption}
\usepackage{subcaption}
\usepackage{algorithm2e}
\usepackage{algpseudocode}

\usepackage{color}
\usepackage{xcolor,colortbl}
\usepackage{framed}
\usepackage{enumitem}

\title{Attentive CutMix: An Enhanced Data Augmentation Approach for Deep Learning Based Image Classification}

\author{%
  Devesh Walawalkar \qquad Zhiqiang Shen\thanks{Corresponding author.} \qquad Zechun Liu \qquad Marios Savvides \\
 Carnegie Mellon University, \\ Department of Electrical and Computer Engineering\\ Pittsburgh, PA, USA
}

\begin{document}

\maketitle

\begin{abstract}
Convolutional neural networks (CNN) are capable of learning robust representation with different regularization methods and activations as convolutional layers are spatially correlated. Based on this property, a large variety of regional dropout strategies have been proposed, such as Cutout~\cite{devries2017improved}, DropBlock~\cite{ghiasi2018dropblock}, CutMix~\cite{yun2019cutmix}, etc. These methods aim to promote the network to generalize better by partially occluding the discriminative parts of objects. However, all of them perform this operation randomly, without capturing the most important region(s) within an object. In this paper, we propose {\em Attentive CutMix}, a naturally enhanced augmentation strategy based on {\em CutMix}~\cite{yun2019cutmix}. In each training iteration, we choose the most descriptive regions based on the intermediate attention maps from a feature extractor, which enables searching for the most discriminative parts in an image. Our proposed method is simple yet effective, easy to implement and can boost the baseline significantly. Extensive experiments on CIFAR-10/100 datasets with various CNN architectures (in a unified setting) demonstrate the effectiveness of our proposed method, which consistently outperforms the baseline {\em CutMix} and other methods by a significant margin.
\end{abstract}
\section{Introduction}
\label{sec:intro}
\begin{figure}[t]
  \centering
  \includegraphics[width=0.6\textwidth]{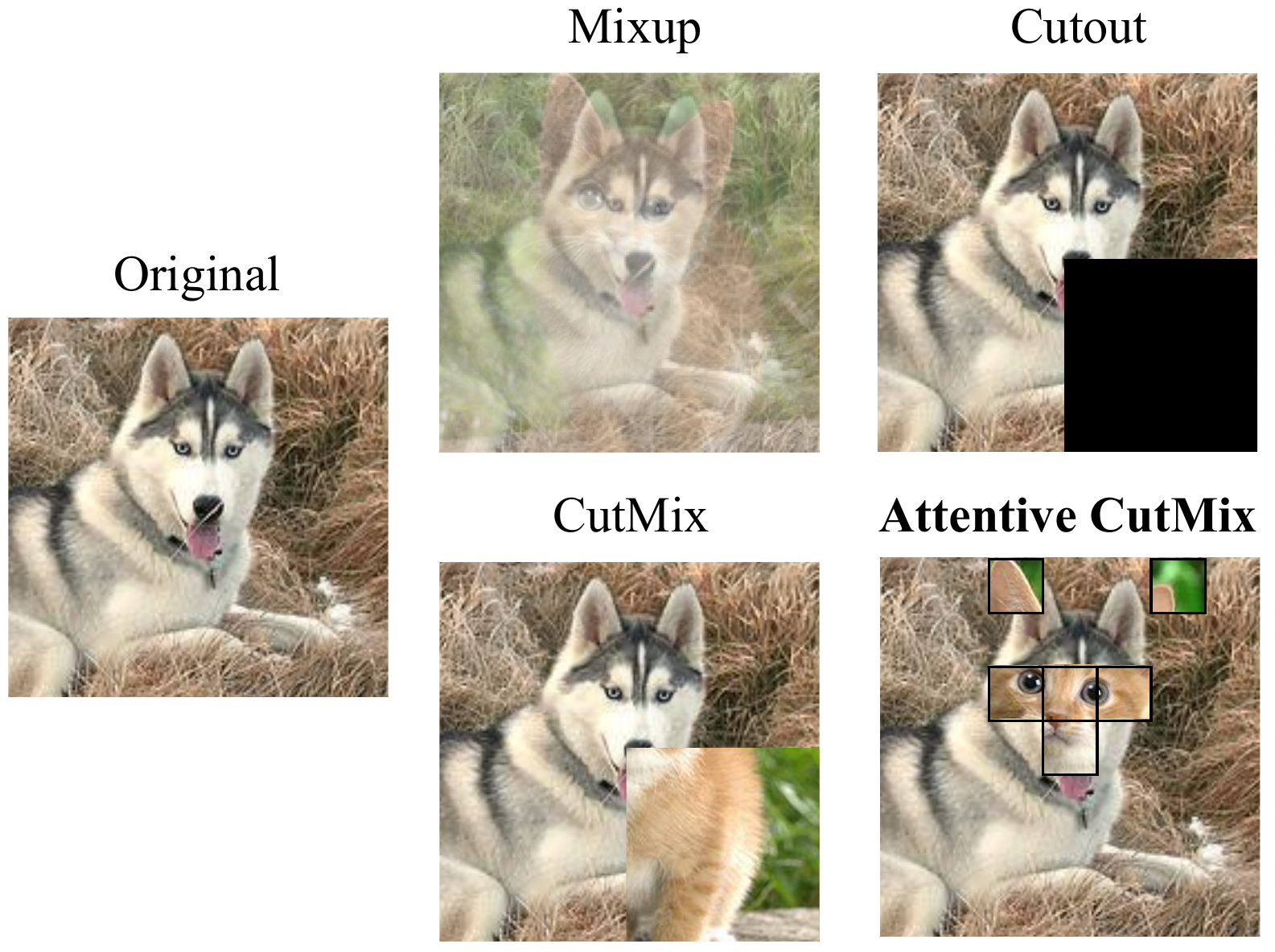}
  \vspace{-0.1in}
  \caption{{Comparison of our proposed {\em Attentive CutMix} with Mixup~\cite{zhang2017mixup}, Cutout~\cite{devries2017improved} and CutMix~\cite{yun2019cutmix}.}}
  \label{fig:comp}
  \vspace{-0.1in}
\end{figure}

Regularization in deep neural networks such as dropout, weight decay, early stopping etc. are popular and effective training strategies to improve the training accuracy, robustness, model performance, while also avoiding overfitting to some extent on limited training data. Among these, dropout is a widely-recognized technique in training neural network which is mainly used in fully connected layers of CNN~\cite{krizhevsky2012imagenet} due to its spatial correlation and dependence property. In recent years, an interesting range of regional dropout or replacement strategies have been proposed, such as Cutout~\cite{devries2017improved}, DropBlock~\cite{ghiasi2018dropblock}, CutMix~\cite{yun2019cutmix}, etc. Specifically, Cutout proposed to randomly mask out square regions of input during training in order to improve the robustness and overall performance of CNNs. DropBlock developed a structured form of dropout that is particularly effective in regularizing CNNs. During training, a contiguous region of a feature map is dropped instead of individual elements in the feature map. CutMix is motivated by Mixup~\cite{zhang2017mixup}, where the regions in an image are randomly cut and pasted among training images and the ground truth labels are also mixed proportional to the area of the regions. Although regional dropout or replacement operation methods have shown great effectiveness of recognition and localization performance in some benchmarks, the dropping or replacing operation is usually randomly conducted on the input. We argue that this strategy may reduce the efficiency of training and also limit the improvement if the networks are unable to capture the most discriminative regions. A representative comparison of our method with other strategies is shown in Fig.~\ref{fig:comp}.

To address the aforementioned shortcoming, we propose to use an {\em attention based CutMix} method. Our goal is to learn a more robust network that can attend to the most important part(s) of an object with better recognition performance without incurring any additional testing costs. We achieve this by utilizing the attention maps generated from a pretrained network to guide the localization operation of cutting and pasting among training image pairs in {\em CutMix}. Our proposed method is extremely simple yet effective. It boosts the strong baseline {\em CutMix} by a significant margin.

We conduct extensive experiments on CIFAR-10/100. Our results show that the proposed {\em Attentive CutMix} consistently improves the accuracy across a variety of popular network architectures (ResNet~\cite{he2016deep}, DenseNet~\cite{huang2017densely} and EfficientNet~\cite{tan2019efficientnet}). For instance, on CIFAR-100, we achieve 75.37\% accurcy with ResNet-152, which is 2.16\% higher than the baseline {\em CutMix} (73.21\%).

\section{Related Work}
\label{sec:relatedwork}

\noindent{\textbf{Data augmentation.}}
Data augmentation operations, such as flipping, rotation, scaling, cropping, contrast, translation, adding gaussian noise etc., are among the most popular techniques for training deep neural networks and improving the generalization capabilities of models. However, in real world, natural data can still exist in a variety of conditions that cannot be accounted for by simple strategies. For instance, the task of identifying the landscape in images can range from rivers, blue sky, freezing tundras to grasslands, forests, etc. Thus, some works~\cite{zhu2017unpaired,huang2018multimodal} have proposed to generate effects such as different seasons artificially to augment the dataset. In this paper, we focus on recognizing natural objects like cat, dog, car, people, etc. wherein we initially discern the most important parts from an object, then use {\em cut} and {\em paste} inspired from {\em CutMix} to generate a new image which helps the networks better attend to the local regions of an image.

\noindent{\textbf{CutMix.}}
{\em CutMix} is an augmentation strategy incorporating region-level replacement. For a pair of images, patches from one image are randomly cut and pasted onto the other image along with the ground truth labels being mixed together proportionally to the area of patches. Conventional regional dropout strategies~\cite{devries2017improved,ghiasi2018dropblock} have shown evidence of boosting the classification and localization performances to a certain degree, while removed regions are replaced usually with zeros or filled with random noise, which greatly reduce/occlude informative pixels in training images. To this end, instead of simply removing pixels, {\em CutMix} replaces the removed regions with a patch from another image, which utilizes the fact that there is no uninformative pixel during training, making it more efficient and effective.

\noindent{\textbf{Attention mechanism.}}
Attention can be viewed as the process of adjusting or allocating activation intensity towards the most informative or useful locations of inputs.
There are several methods for exploiting attention mechanism to improve image classification~\cite{wang2017residual,hu2018squeeze} and object detection~\cite{shenimproving} tasks. GFR-DSOD~\cite{shenimproving} proposed a gated CNN for object detection based on \cite{shen2017dsod} that passes messages between features from different resolutions and used gated functions to control information flow. SENet~\cite{hu2018squeeze} used attention mechanism to model channel-wise relationships and enhanced the representation ability of modules through
the networks. In this paper, we introduce a simple attention-based region selection that can find out the most discriminative parts spatially.

 \begin{figure*}[t]
  \centering
  \includegraphics[width=0.9\textwidth]{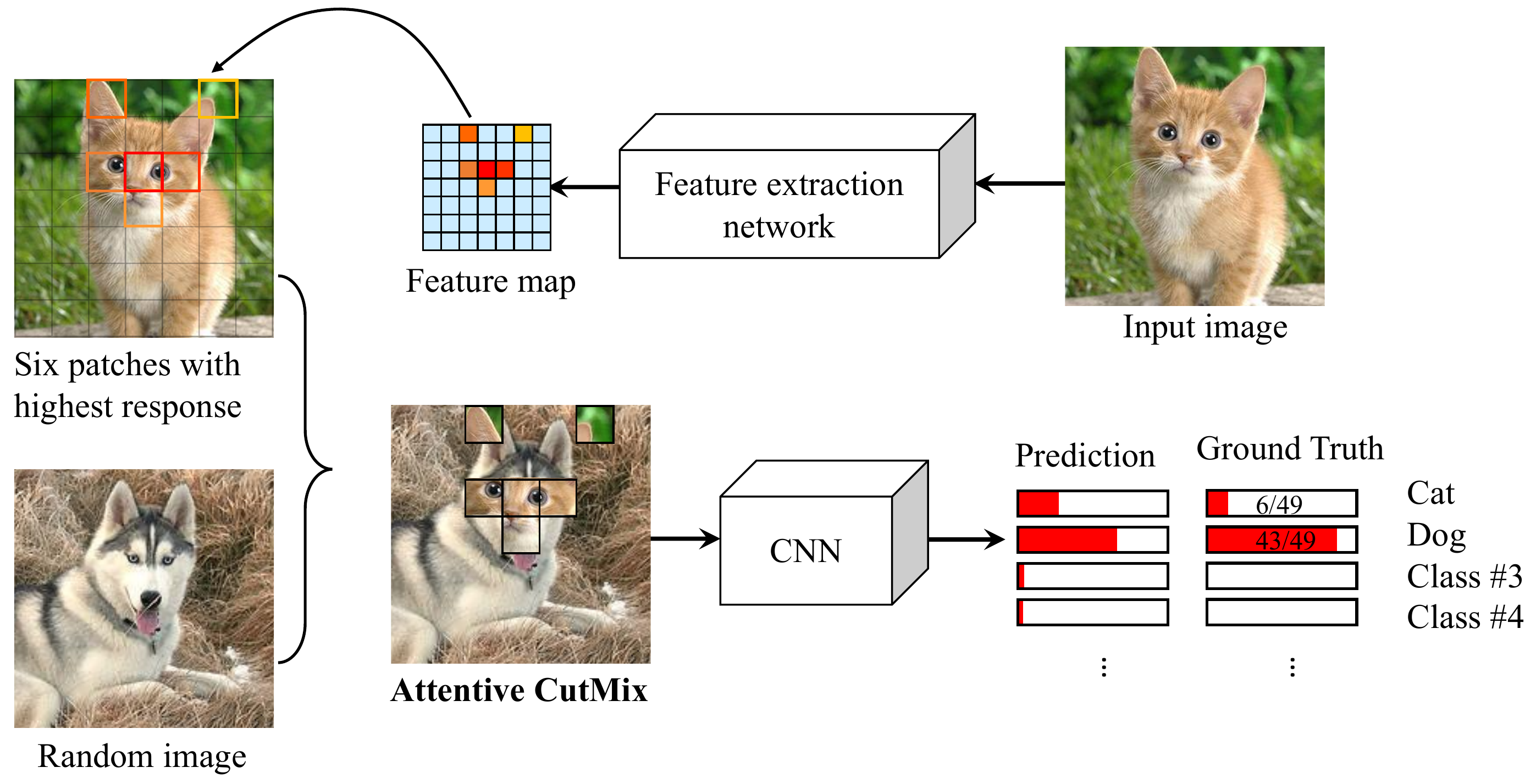}
  \vspace{-0.1in}
  \caption{Framework overview of proposed {\em Attentive CutMix}.}
  \label{fig:framework}
  \vspace{-0.1in}
\end{figure*}

\section{Proposed Approach}
\label{sec:approach}
\subsection{Algorithm}
The central idea of {\em Attentive CutMix} is to create a new training sample $(\Tilde{x},\Tilde{y})$ given two distinct training samples $(x_1,y_1)$ and $(x_2,y_2)$. Here, $x \in R^{W \times H \times C}$ is the training image and $y$ is the training label. Similar to CutMix \cite{yun2019cutmix}, we define this combining operation as,
\begin{equation*}
    \Tilde{x} = \textbf{B} \odot x_{1} + (\textbf{1} - \textbf{B}) \odot x_{2}
\end{equation*}
\begin{equation}
    \Tilde{y} = \lambda y_{1} + (1 - \lambda) y_{2}
\end{equation}
    where $\textbf{B} \in \{0,1\}^{W \times H}$ denotes a binary mask indicating which pixels belong to either of the two images, \textbf{1} is a binary mask filled with ones and $\odot$ is the general element-wise multiplication. Here $\lambda$ is the ratio of patches cut from the first image and pasted onto the second image to the total number of patches in the second image.\\
    We first obtain a heatmap (generally a 7$\times$7 grid map) of the first image by passing it through a pretrained classification model like ResNet-50 and take out the final 7$\times$7 output feature map. We then select the top ``$N$'' patches from this 7$\times$7 grid as our attentive region patches to cut from the given image. Here $N$ can range from 1 to 49 (i.e. the entire image itself). Later, we will present an ablation study on the number of attentive patches to be cut from a given image.
    
    We then map the selected attentive patches back to the original image. For example, a single patch in a 7$\times$7 grid would map back to a 32$\times$32 image patch on a 224$\times$224 size input image. The patches are cut from the first image and pasted onto the second image at their respective original locations, assuming both images are of  the same size. The pair of training samples are randomly selected during each training phase. For the composite label, considering that we pick the top 6 attentive patches from a 7$\times$7 grid, $\lambda$ would then be $\frac{6}{49}$. Every image in the training batch is augmented with patches cutout from another randomly selected image in the original batch. Please refer Fig.~\ref{fig:framework} for an illustrative representation of our method.

\subsection{Theoretical Improvements over CutMix}
   CutMix provides good empirical capabilities of improving the classification accuracy of Deep Learning models. However, there are weak theoretical foundations to its effectiveness. One of the reasons for its effectiveness could be that pasting random patches onto an image provides random occlusions to the main classification subject in the image thus making it harder for the model to overfit on a particular subject and forces it to learn more important features associated with a given subject. 
   
   However, the patch cutout is of random size and taken from a random location, thus creating the possibility of cutting an unimportant background patch and simultaneously pasting it onto the background in the second image. Since the composite label contains a part of first label, we are theoretically associating the background region to that label for the model to learn. This hinders the empirical gains of CutMix and this is where {\em Attentive CutMix} provides improvements over its CutMix counterpart.
   
   Rather than randomly selecting the patch, {\em Attentive CutMix} takes help of a pretrained network to determine the most important or representative regions within the image for the classification task. This technique's effectiveness thus directly co-relates to the pretrained model used. The better the model the more effective {\em Attentive CutMix} will be. Also, the cutout attentive patch is pasted onto the same area in the second image as it was in the original image. This further helps to better occlude the image since the pasting randomization in CutMix does provide a possibility of the patch being pasted onto the background rather than the object of interest. Thus {\em Attentive CutMix} improves on dual fronts of patch selection and pasting by removing the randomness and using attentive intuition to make more robust fusing of images.


\section{Experiments and Analysis}
\label{sec:ex[s}

\subsection{Datasets and Models}
To prove the effectiveness of our data augmentation technique we perform extensive experiments across wide range of popular models and Image classification datasets. We select four variants each of ResNet \cite{he2016deep}, DenseNet~\cite{huang2017densely} and EfficientNet~\cite{tan2019efficientnet} architectures. We selected these particular architecture since they provide substantial variation in their architectural concepts. The individual variants in each architecture help us test the method across different depths/sizes of each architecture. For image classification datasets we chose CIFAR-10 and CIFAR-100 \cite{krizhevsky2009learning} as these are widely used benchmark datasets to compare our method against.   

\subsection{Implementation Details}
 We trained the individual models from scratch (except the efficientnet which is too costly to train from scratch) to prevent any pretraining bias to affect our evaluation results. We run the baselines according to the hyperparameter configurations used in their original papers,  
 but due to some absence of implementation details, on some particular datasets/networks the settings may be slightly different. 
 All the data augmentation techniques for a given architecture and dataset were run for a fixed number of epochs which were enough for the models to converge onto a test set accuracy, considering our prime objective was to test other data augmentation techniques against ours rather than matching state-of-the-art results. All the models were implemented in Pytorch~\cite{paszke2017automatic} framework and all data augmentation technique implementation was also done in requirements with the Pytorch framework. 
 
\subsection{Results on CIFAR-10}
For CIFAR-10 dataset, each of the models for a given data augmentation technique was trained for 80 epochs. The batch size was kept at 32 and the learning rate at 1e-3. The model weights were initialized using Xavier Normal technique. Weight decay was incorporated for regularization and it's value was kept at 1e-5. Results are presented in Table \ref{tab:res_c10}. Our method provides better results over all tested models compared to CutMix, Mixup and the baseline methods.

\begin{table}[h]
\centering
\resizebox{.75\textwidth}{!}{%
\begin{tabular}{l|c|c|c|c}
\Xhline{1pt}
\hline
           & \multicolumn{4}{c}{\bf CIFAR-10 (\%)}                \\ \hline
Method     & \bf Baseline & \bf Mixup & \bf CutMix & \bf Attentive CutMix \\ \hline
ResNet-18  & 84.67    & 88.52 & 87.92   &\bf 88.94            \\ 
ResNet-34  & 87.12    & 88.70 & 88.75  & \bf 90.40            \\ 
ResNet-101 &90.47     & 91.89 & 92.13  & \bf 93.25               \\ 
ResNet-152 & 92.45    & 94.21 & 94.35 & \bf 94.79                 \\ \hline
DenseNet-121 & 85.65  & 87.56 & 87.98 & \bf 88.34              \\ 
DenseNet-169 & 87.67 &  89.12 & 89.23 & \bf 90.45               \\ 
DenseNet-201 & 91.21 &  93.21 & 93.45 & \bf 94.16             \\ 
DenseNet-264 & 92.78 &  94.20 & 94.34 & \bf 94.83             \\ \hline
EfficientNet - B0 & 87.45 &  88.07 & 88.67 &  \bf 88.94              \\ 
EfficientNet - B1 & 90.12 &   90.99  &  91.37 & \bf 92.10               \\ 
EfficientNet - B6 & 92.74 &  93.76 &  93.28 & \bf 93.92             \\ 
EfficientNet - B7 & 94.95 &  95.11 &  95.25 & \bf 95.86            \\ \hline \Xhline{1pt}
\end{tabular}}
\vspace{0.1in}
\caption{Comparison of accuracy (\%) with {\em baseline}, {\em Mixup} and {\em CutMix} on CIFAR-10.}
\label{tab:res_c10}
\vspace{0.1in}

\centering
\resizebox{.75\textwidth}{!}{%
\begin{tabular}{l|c|c|c|c}
\Xhline{1pt}
\hline
           & \multicolumn{4}{c}{\bf CIFAR-100 (\%)}               \\ \hline
Method     & \bf Baseline & \bf Mixup & \bf CutMix & \bf Attentive CutMix \\ \hline
ResNet-18  & 63.14    & 64.40 &   65.90   & \bf 67.16            \\ 
ResNet-34  & 65.54    & 67.83 &   68.32   & \bf 70.03            \\ 
ResNet-101 &68.24     & 70.76 &  71.32    & \bf 72.86               \\ 
ResNet-152 & 71.49    & 74.81 &  73.21    & \bf 75.37            \\ \hline
DenseNet-121 &65.12  &   66.84  & 67.62 & \bf 69.23            \\ 
DenseNet-169 & 66.42 &   68.24  & 69.58 & \bf 71.34            \\ 
DenseNet-201 & 70.28 &  72.89 & 73.57 & \bf 74.65              \\ 
DenseNet-264 & 73.51 & 76.49  & 75.23 & \bf 77.58                 \\  \hline
EfficientNet - B0 & 64.67 &  65.78 & 66.95 & \bf 67.48              \\ 
EfficientNet - B1 &  66.89 &  68.23 & 68.12 & \bf 68.96              \\ 
EfficientNet - B6 & 71.34 &  73.56 & 73.75 & \bf 74.82             \\ 
EfficientNet - B7 & 75.67 &  77.21 & 77.57 & \bf 78.52            \\ \hline \Xhline{1pt}
\end{tabular}}
\vspace{0.1in}
\caption{Comparison of accuracy (\%) with {\em baseline}, {\em Mixup} and {\em CutMix} on CIFAR-100.}
\label{tab:res_c100}
\end{table}

\subsection{Results on CIFAR-100}
For CIFAR-100 dataset, every model was trained for 120 epochs. The batch size was kept at 32 and the learning rate at 1e-3. The model weights were initialized using Xavier Normal technique. Weight decay was incorporated for regularization and it's value was kept at 1e-5. Our experiments are presented in Table \ref{tab:res_c100}. Our method again provides better overall results compared to other data augmentation methods.

\subsection{Ablation Study}
The number of patches ``$N$'' to be cutout from the first image is a hyperparameter that needs to be tuned for optimal performance of our method. We conducted a study for the optimal value of this hyperparameter across the range of 1 to 15. We tested each value in this range for all our experiments and found out that cutting out top 6 attentive patches gave us the best average performance across the experiments. One of the explanations for this value could be that cutting out less than 6 patches doesn't provide enough occlusion to the main subject in the second image. On the contrary, cutting more than 6 patches might be providing excessive occlusion to the original subject in the image so as to make the respective label for that image not enough discriminative for the model to learn anything useful.   

\vspace{-0.1in}
\subsection{Discussion}
Our experiments do provide strong evidence that our method provides much better results than CutMix and Mixup on average over different datasets and architectures. {\em Attentive CutMix} consistently provides an average increase of 1.5\% over other methods which validates the effectiveness of our attention mechanism. One disadvantage of our method is the fact that a pretrained feature extractor is needed in addition to the actual network that is to be trained. However, depending on the classification task and training complexity of the model and dataset used, we can vary the size of pretrained extractor being used in data augmentation. Thus this can prove be a minor extra computation in the overall training scheme and it can easily be offset by the performance gains of this method. We believe that this method could provide similar gains for other computer vision tasks like object detection, instance segmentation, etc., since all rely on robust features being extracted from the image, which is also the main foundation for image classification task. The future direction for this work would thus be to incorporate this method for other vision tasks.    

\section{Conclusion}
\label{sec:conclusion}
We have presented {\em Attentive CutMix}, a attention-based data augmentation method that can automatically find the most discriminative parts of an object and replace them with patches cutout from other image. Our proposed method is simple yet effective, straight-forward to implement and can help boost the baseline significantly. Our experimental evaluation on two benchmarks CIFAR-10/100 verifies the effectiveness of our proposed method, which obtains consistent improvement for a variety of network architectures.

 \bibliographystyle{IEEEbib}
 \bibliography{Main}


\end{document}